# Blinded Multi-Rater Comparative Evaluation of a Large Language Model and Clinician-Authored Responses in CGM-Informed Diabetes Counseling

Zhijun Guo[1]; Alvina Lai[1]; Emmanouil Korakas[2]; Aristeidis Vagenas[2]; Irshad Ahamed[3]; Christo Albor[4]; Hengrui Zhang[1]; Justin Healy[4]; Kezhi Li[1,2]

[1] Institute of Health Informatics, University College London, London, United Kingdom

[2] University College London Hospitals NHS Foundation Trust, London, United Kingdom

[3] Dartford and Gravesham NHS Foundation Trust, Dartford, United Kingdom

[4] Royal Free London NHS Foundation Trust, London, United Kingdom

**Corresponding author:**

Kezhi Li

Institute of Health Informatics

University College London

222 Euston Road

London

United Kingdom

Phone: +44 7859 995590

Email: ken.li@ucl.ac.uk

## Abstract

**Background:**

Continuous glucose monitoring (CGM) is central to modern diabetes care, but explaining CGM patterns clearly, consistently, and empathetically remains time-intensive in practice. Large language model (LLM)–based systems may support patient-facing interpretation of CGM data, but evidence remains limited for retrieval-grounded tools evaluated against clinician-authored responses in counseling scenarios. The system was intended for structured CGM interpretation and communication support rather than autonomous therapeutic decision making.

**Objective:**

To evaluate whether a retrieval-grounded LLM-based conversational agent (CA) could support patient understanding of CGM data and preparation for routine diabetes consultations by generating responses to questions arising during CGM-informed diabetes counseling, with quality comparable to clinician-authored responses.

**Methods:**

We developed a retrieval-grounded LLM-based CA for CGM interpretation and diabetes counseling support. The system was designed to provide plain-language explanations of CGM patterns and responses to diabetes management questions while avoiding directive or individualized medical advice, such as recommending medication initiation, dose adjustment, or regimen changes. 12 CGM-informed cases, each comprising a de-identified CGM trace, a synthetic patient vignette, and accompanying CGM visual materials, were constructed from publicly available clinical datasets. Between Oct 2025 and Feb 2026, six senior UK diabetes clinicians each reviewed 2 assigned cases and answered 24 questions (12 per case). In a blinded multi-rater evaluation, each CA-generated and clinician-authored response was independently rated by 3 clinicians on 6 quality dimensions: clinical accuracy, guideline adherence, actionability, personalization, communication clarity, and empathy. Safety flags and perceived source labels were also recorded. The primary analysis used linear mixed-effects models with random intercepts for case and rater.

**Results:**

A total of 288 unique responses (144 CA and 144 clinician responses) were evaluated, generating 864 ratings. The CA received higher quality scores than clinician responses (mean 4.37 vs 3.58), with an estimated mean difference of 0.782 points on a 5-point scale (95% CI 0.692-0.872; P<.001). This pattern was observed across all 6 categories of patient questions. The largest estimated differences were for empathy (mean difference 1.062, 95% CI 0.948-1.177) and actionability (0.992, 95% CI 0.877-1.106). Safety flag distributions were similar between CA and clinician responses, with major concerns rare in both groups (3/432, 0.7% each). Although CA responses were longer, additional analyses adjusting for word count did not indicate that response length explained the overall quality difference.

**Conclusions:**

Retrieval-grounded LLM-based systems may have value as adjunct tools for routine CGM review, patient education, and preconsultation preparation, with potential to reduce clinician time spent on standardized interpretive tasks. However, these findings should be interpreted in light of the vignette-based design, restricted datasets, and a small clinician panel, and they do not establish suitability for autonomous therapeutic decision-making, medication adjustment, or unsupervised real-world use. Prospective validation in interactive clinical workflows is needed before implementation.

# Introduction

**Background**

Diabetes mellitus is a chronic metabolic disease characterized by hyperglycemia due to impaired insulin secretion, insulin action, or both, and is associated with substantial microvascular and macrovascular complications [1]. The International Diabetes Federation (IDF) reports that in 2024, approximately 589 million adults aged 20–79 years were living with diabetes worldwide (11.1% of the adult population), with this number projected to reach 853 million by 2050, an increase of 46% [2]. Given this growing global burden, long-term diabetes management increasingly depends on tools that can support day-to-day monitoring, interpretation of glycemic patterns, and communication around self-management [3,4].

Continuous glucose monitoring (CGM) has become central to contemporary diabetes care [5]. CGM systems provide near real-time interstitial glucose readings and trend information, enabling assessment of time in range, glycemic variability, and hypoglycemia risk [6,7]. Evidence from clinical trials and consensus reports indicates that CGM use can increase time in range, reduce hypoglycemia, and improve treatment satisfaction in both type 1 and type 2 diabetes [6,8]. Recent American Diabetes Association (ADA) Standards of Care recommend CGM for most individuals on intensive insulin therapy and support broader adoption where feasible [1,9]. As device accuracy, wearability, and reimbursement have improved, CGM use in routine clinical practice has expanded rapidly [9,10].

Despite these benefits, interpreting CGM data remains challenging for many people living with diabetes. Although uptake is increasing, access and sustained use remain affected by device cost, insurance coverage, and variation in prescribing practices across clinical settings [11]. Modern CGM platforms increasingly provide structured summaries of glucose data for patient and clinical review, including pattern reports, trend visualizations, and summary statistics [12,13]. Dexcom Clarity, for example, highlights glucose patterns, trends, and statistics through report-based summaries [12], whereas Abbott's LibreView provides reports such as Glucose Pattern Insights that highlight glycemic patterns and may include medication and lifestyle considerations for review [13]. However, these tools are primarily designed to summarize and visualize data rather than to provide conversational, patient-facing explanations of why patterns may be occurring or how they relate to daily behaviors, treatment routines, and patient concerns. Even among CGM users, limited structured training in how to interpret traces means that many still feel overwhelmed by the volume and complexity of CGM readouts and struggle to relate observed patterns to food intake, physical activity, medication timing, or other day-to-day behaviors [14]. Patients may be unsure how to respond to recurring highs or lows, how to understand trends over time, or which questions should be brought to clinicians during review [15-17]. Limited consultation time further constrains opportunities for detailed, individualized explanation of CGM traces in standard clinical encounters [14,15]. Together, these practical and educational barriers highlight a need for scalable approaches that can support clear, timely, and patient-facing interpretation of CGM data alongside routine care.

In addition to these interpretive challenges, many people living with diabetes experience substantial emotional burden. Meta-analyses suggest that approximately 10%–15% of adults with diabetes have clinically significant depressive symptoms [18]. Diabetes distress and depression are associated with poorer glycemic control, reduced adherence to self-management behaviors, and lower quality of life [19]. In this context, support tools for CGM-related communication must address not only the technical explanation of glucose patterns but also the uncertainty, frustration, and anxiety that often accompany diabetes self-management. This does not necessarily require formal psychological intervention, but it does underscore the importance of clarity, reassurance, and empathic communication in patient-facing diabetes support.

Recent advances in large language models (LLMs) offer a potential way to address unmet informational and communication needs in diabetes care [20]. LLMs can accept structured and unstructured inputs, including numerical summaries, clinical text, and free-text patient queries, and generate contextualized natural-language explanations [21]. This makes them candidates for supporting tasks such as CGM interpretation, patient education, and communication of case-specific information in accessible language [22,23]. Early evaluations and review articles report that, in many scenarios, LLM-based systems provide reasonably accurate health information and can generate responses that show elements of cognitive empathy, such as recognizing expressed emotions and offering supportive language in simulated patient interactions [22-25]. However, current LLMs in medicine remain limited by hallucinated or incomplete content, variable alignment with clinical guidelines and challenges in transparently incorporating patient-specific structured data, which raises concerns about their safe use in clinical contexts [22,26]. Retrieval-augmented generation (RAG) has been proposed as one strategy to mitigate these limitations by grounding LLM outputs in curated, evidence-based sources and explicitly linking responses to both guideline documents and case-specific data such as CGM traces [25,26]. Empirical evidence on the performance of such systems in patient-specific, CGM-informed diabetes scenarios, however, remains limited.

Recent studies have applied LLM- or RAG-based conversational agents (CAs) to diabetes care, mainly for diabetes education, lifestyle and self-management advice, or general question answering [27-28]. These systems are commonly evaluated in terms of diabetes knowledge, perceived usefulness, usability, and the appropriateness or safety of chatbot advice [27-30], but they rarely incorporate CGM data directly or examine how such systems perform when asked to interpret CGM-informed cases in a patient-facing manner. In parallel, other studies have used LLMs to analyze CGM data by calculating standard CGM metrics and generating narrative summaries of glucose traces, which endocrinologists then assess for accuracy, completeness, and safety, often using simulator-generated rather than real-world data [31]. Collectively, these studies suggest that LLM-based tools may assist with diabetes-related information provision and CGM analysis [27-31]. However, they have not adequately evaluated retrieval-grounded LLM-based CAs in CGM-informed scenarios that require structured glucose interpretation together with patient-centered communication, nor have they compared specialist ratings of CA-generated responses against clinician-authored responses to the same cases.

In practice, one plausible role for such a system may be as an adjunct to routine diabetes care, particularly for preconsultation preparation, standardized explanation of CGM patterns, and support for common patient-facing questions that are structured but time-consuming to address repeatedly in routine care [32,33]. Before a scheduled review, patients might use such a system to obtain a plain-language explanation of recent glucose patterns, identify recurring concerns, and formulate questions for discussion with their diabetes care team. In this way, the system could potentially help surface common issues in advance, support more consistent handling of routine CGM-related questions, and improve consultation efficiency by allowing clinicians to focus more directly on individualized decision making and more complex cases. More broadly, this suggests a possible practical role for LLM-based tools in diabetes services, where many routine advisory tasks involve synthesizing structured and semistructured information into clear, context-specific communication, while final treatment recommendations and higher-risk clinical judgments remain clinician-led. Accordingly, the potential value of such systems may lie less in autonomous decision making than in supporting routine explanatory work, augmenting specialist capacity, and helping prioritize clinician time toward situations requiring greater clinical complexity or safety oversight [33,34].

### Study Aim

To our knowledge, this study is among the first to conduct a blinded comparative evaluation in which diabetes specialists assessed both retrieval-grounded LLM-generated and clinician-authored responses to identical CGM-informed cases across predefined clinical and psychosocial evaluation criteria. To address the need for scalable support in patient-facing communication around CGM review, we developed a retrieval-grounded LLM (GPT-5.1)-based CA that generated structured, plain-language responses to questions arising in CGM-informed diabetes counseling using case-specific CGM information, vignette context, and guideline-based retrieval. We then conducted a blinded multi-rater evaluation between Oct 2025 and Feb 2026 with UK diabetes clinicians, who assessed CA- and clinician-authored responses for quality and safety. In doing so, this study examined whether such a system could support standardized explanatory and educational aspects of diabetes care in consultations involving CGM review, without positioning it as a substitute for clinician judgment.

## Methods

### Ethical Considerations

This study did not involve direct contact with patients or access to identifiable clinical records. The CGM files were derived from publicly available, preexisting, de-identified clinical datasets, and all accompanying patient vignettes were synthetic and created solely for research purposes. The human contributors were 6 diabetes clinicians who served as expert reviewers by providing written responses and rating anonymized text outputs. They were not asked to provide personal or sensitive information and were considered expert assessors rather than research participants. According to UK Health Research Authority guidance, research using publicly available or fully anonymized data in which individuals cannot be identified may not require formal Research Ethics Committee review [35]. Because the present study used publicly available anonymized CGM datasets together with clinician ratings of generated responses and did not involve identifiable patient information, formal ethics approval was not required.

### CGM Data and Case Vignettes

An overview of the complete workflow is presented in Figure 1. We constructed 12 CGM-based cases using de-identified glucose traces from two publicly available research datasets. Six traces (3 type 1 diabetes and 3 type 2 diabetes) were sampled from the ShanghaiT1DM and ShanghaiT2DM datasets [36], which provide 3–14 days of CGM values at 15-minute intervals for 12 individuals with type 1 diabetes and 100 individuals with type 2 diabetes, together with capillary blood glucose measurements, blood ketone, self-reported dietary intake, insulin doses, and clinical characteristics [36]. For the 6 Shanghai cases used in this study, the CGM monitoring period ranged from 7 to 14 days: 1 case included 7 consecutive days of data, and the remaining 5 included at least 10 days of CGM readings. In addition to the CGM glucose time series, selected contextual fields from the source datasets, including dietary intake and insulin records, were used to inform construction of the synthetic patient vignettes. These variables were used only for case contextualization and were not analyzed as separate study variables.

The remaining 6 traces were selected from the OhioT1DM 2018 dataset [37], which contains 8 weeks of CGM, insulin pump, physiological sensor, and self-reported life-event data for 6 adults with type 1 diabetes [37]. In addition to the CGM glucose measurements, selected self-reported life-event information was used to inform the construction of the synthetic patient vignettes. As with the Shanghai cases, these variables were used only for case contextualization and were not analyzed as separate study variables. The resulting case set comprised 9 type 1 diabetes cases and 3 type 2 diabetes cases. This distribution reflected both the composition of the source datasets and the more limited availability of suitable publicly available type 2 diabetes CGM datasets for case construction, while allowing the evaluation to include both major diabetes types and a range of CGM pattern presentations. Within the type 2 diabetes group, cases were not further selected to represent specific treatment regimens, such as insulin-treated or non-insulin-treated subgroups. Instead, the selected cases should be interpreted as examples of CGM-informed counseling scenarios in type 2 diabetes rather than as representing the full clinical heterogeneity of the broader type 2 diabetes population.

To approximate a realistic consultation context while preserving privacy, we created a synthetic case vignette for each selected CGM trace. Each vignette included a brief clinical profile (for example, age band, sex, diabetes type and duration, treatment regimen, typical diet and activity patterns, and the patient's main concerns about glycemic control) and one or more CGM plots generated from the underlying trace to give clinicians an intuitive visual overview of glucose patterns (example can be found in Multimedia Appendix 1). Draft vignettes and plots were prepared by Z.G and refined in two rounds of review with senior diabetologist E.K to ensure clinical plausibility, consistency with the CGM data, and absence of identifiable details.

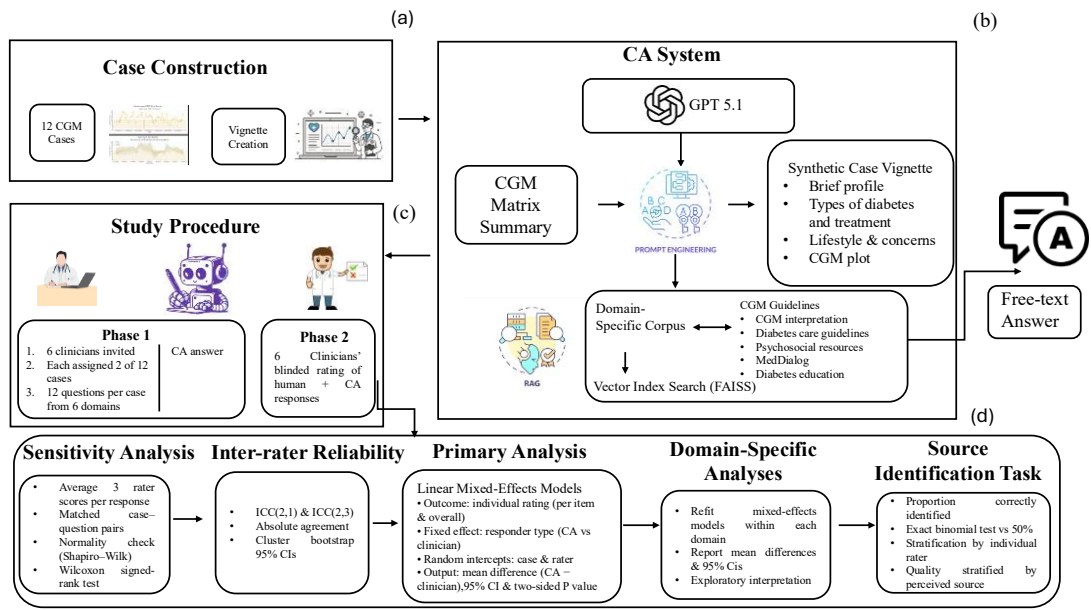

**Figure 1: Overview of the study workflow, including case construction, CA system design, study procedures, and statistical analyses.**

*(a) Case Construction:* 12 de-identified CGM cases were created using time series data and corresponding synthetic vignettes describing diabetes type, treatment regimen, lifestyle patterns, and glycemic concerns.

*(b) CA System:* The CA was implemented using GPT-5.1 with prompt-based adaptation. Each model input consisted of (1) a CGM metrics summary generated from Python preprocessing, (2) the synthetic vignette, and (3) relevant guideline excerpts retrieved through RAG using a domain-specific corpus indexed in FAISS. The structured system prompt incorporated safety and empathy instructions. The model produced one free-text answer per case-question pair.

*(c) Study Procedure:* 6 clinicians participated in two phases. In Phase 1, each clinician answered questions for two assigned cases while the CA generated parallel responses. In Phase 2, clinicians performed blinded ratings of responses written by other clinicians and the CA based on predefined quality dimensions.

*(d) Statistical Analysis:* Analyses included inter-rater reliability assessment using two-way random-effects intraclass correlation coefficients (ICCs), primary linear mixed-effects models comparing clinician and CA ratings, sensitivity analyses using paired Wilcoxon signed-rank tests, domain-specific mixed-effects models, and source-identification analyses using exact binomial tests.

## CA Design

A retrieval-grounded CA for CGM interpretation and diabetes-related counseling scenarios was implemented using the GPT-5.1 LLM accessed via the OpenAI application programming interface (API). GPT-5.1 was one of the most advanced models available at the time of system development; no comparative benchmarking across alternative models was undertaken. The base model was used without parameter fine-tuning. Task adaptation relied on structured prompting combined with RAG. De-identified CGM summaries and vignette-based patient queries were embedded into predefined prompt templates, and guideline-aligned reference materials were retrieved from a curated knowledge base to condition response generation.

For each case, the CGM glucose time series was processed in Python using a custom analysis function that calculated core consensus metrics, including monitoring duration, data completeness, mean glucose, SD, coefficient of variation, and the percentage of readings within standard CGM ranges (time in range [TIR] 70–180 mg/dL [3.9–10.0 mmol/L], time below range [TBR] <70 mg/dL [<3.9 mmol/L] and <54 mg/dL [<3.0 mmol/L], and time above range [TAR] >180 mg/dL [>10.0 mmol/L] and >250 mg/dL [>13.9 mmol/L]), consistent with international recommendations for CGM-derived metrics and TIR reporting [6]. The function generated a concise text summary listing these metrics alongside commonly used target thresholds. This standardized CGM summary was used as structured input to the CA.

To verify the correctness of the implementation, Z.G manually recalculated the CGM metrics for representative traces from both the Shanghai and Ohio datasets using Microsoft Excel as an implementation check. For these validation traces, all metrics were recomputed using simple count and average formulas based on the international CGM TIR consensus definitions [38], and the Excel results exactly matched the Python outputs at the precision reported. This validation step was performed solely to confirm the correctness of the implementation and to support basic quality checks, such as visual inspection that the reported metrics were consistent with the plotted CGM traces. The metric calculations themselves were not shown to clinician raters. For each question, the model input consisted of the CGM summary, the synthetic case vignette (clinical profile and contextual information), the text of a pre-specified question from the case-specific question set, and image-based CGM visualizations for the case, including both the continuous glucose trace and an ambulatory glucose profile (AGP)-style profile. These visualizations were supplied directly to the CA as multimodal inputs, so the model's interpretation was informed by both structured summary metrics and visual glucose pattern information. An example CGM summary format is provided in Multimedia Appendix 1.

To ground responses in current evidence and professional guidance, a RAG approach was used. A domain-specific corpus was built from publicly available materials on CGM metrics and interpretation [38-41], CGM-guided glucose management [42-46], AGP reports [39], major diabetes care guidelines [47,48], and resources on emotional and psychosocial aspects of diabetes [47-51]. The corpus also included selected segments from the MedDialog 2020 medical dialogue dataset to provide examples of clinician–patient conversational style [52], and curated patient education materials from national health services and diabetes charities (such as Diabetes UK [50] and the ADA [51]), which were manually organized into topic-based summaries. All documents were converted to plain text and split into short overlapping segments using a recursive character-based splitter with a target segment length of approximately 500 characters and an overlap of 100 characters. Each segment was embedded using OpenAI's text embedding service via the LangChain OpenAIEmbeddings wrapper [53], and stored in a FAISS (Facebook AI Similarity Search) vector index implementing approximate nearest-neighbour search [54]. At inference time, the text of the current question was used as the retrieval query to obtain the top two most similar segments from the vector index based on the default cosine-like similarity measure. These retrieved segments were then inserted into the prompt as reference material, together with the CGM summary and vignette for the model's answer.

Two prompt templates were used to structure model behavior across interaction stages. The first template was used to generate an initial plain-language explanation of the precomputed CGM summary and vignette context, with retrieved guideline-aligned reference materials incorporated through the RAG pipeline to support interpretation. The CGM metrics themselves were computed in Python in advance and provided to the model as structured input rather than being generated by the prompt. The second template was used to answer follow-up patient questions, again incorporating retrieved reference materials through the RAG pipeline to support context-aware responses. Full prompt templates are provided in Multimedia Appendix 2. The system prompt defined the CA as a diabetes specialist and instructed it to provide clear, plain-language, empathetic, and non-judgmental explanations aligned with standard diabetes education practices. During system development, selected example outputs were reviewed by a clinical collaborator (E.K) to provide targeted feedback on tone, clarity, and clinical appropriateness, and this feedback informed the final prompt refinement. For each case–question pair in the formal evaluation, however, the model produced a single free-text response without iterative refinement.

### Safety and Security

Several safeguards were implemented to minimize the risk of inappropriate, misleading, or clinically unsafe outputs. Inputs to the CA were restricted to de-identified CGM summaries and synthetic vignettes, so no identifiable patient information or directly actionable treatment instructions were supplied. Behavioral constraints embedded in the system prompted the CA to avoid prescriptive therapeutic recommendations and to encourage consultation with the person's usual diabetes care team when glycemic patterns suggested potential risk, reinforcing its role as a communication support tool for structured CGM explanation rather than a substitute for clinical judgment. The RAG component grounded responses in a curated, guideline-consistent corpus rather than unrestricted external sources, and all CGM metrics used to contextualize model reasoning were validated in advance to avoid propagating incorrect numerical information. During system development, draft outputs, for example, case–question pairs were iteratively reviewed by diabetes clinicians, and their feedback was used to refine the prompt and guardrails until the response style was judged acceptable for the study.

## Study Design

This study used a vignette-based, multi-rater, blinded comparative design to evaluate the CA's responses against those of diabetes clinicians. The expert panel comprised 6 senior diabetes specialists recruited from major UK National Health Service (NHS) Foundation Trusts, including Imperial College Healthcare NHS Trust (St Mary's Hospital), University College London Hospitals NHS Foundation Trust (UCLH), and the Royal Free London NHS Foundation Trust. The use of senior diabetes specialists provided a stringent benchmark for comparison.

Participants met pre-defined eligibility criteria: (1) a primary clinical appointment within a tertiary referral center or major teaching hospital; (2) at least 10 years of post-qualification clinical experience; and (3) demonstrated expertise in diabetes technology, particularly in the clinical interpretation of CGM and advanced insulin delivery systems. The panel included consultant diabetologists and senior specialists, several of whom hold clinical-academic appointments.

After constructing 12 CGM-based cases with synthetic clinical vignettes as described above, 6 UK diabetes clinicians were first invited to provide written responses and subsequently to rate anonymized response pairs. Each clinician was randomly assigned 2 of the 12 cases and received the full standardized case package for each assigned case. This included the synthetic vignette together with structured supporting materials, including demographic and clinical background, current treatment, CGM summary metrics, lifestyle and self-management information, patient-reported observations, accompanying CGM visual materials, and the raw CGM data.

The question set was grouped into 6 content domains reflecting topics commonly discussed in diabetes consultations in which CGM review forms an important component: (A) blood glucose interpretation and fluctuation analysis (6 questions), (B) impact of food and exercise (7 questions), (C) medication and treatment guidance (4 questions), (D) emotional and psychological concerns (5 questions), (E) long-term goals and motivation (5 questions), and (F) technical issues and device use (4 questions). These domains were selected to reflect the breadth of patient-facing questions that may arise in such consultations, ranging from direct interpretation of glucose patterns to treatment-related, psychosocial, goal-oriented, and device-related concerns [55]. The draft question bank was reviewed by senior diabetologist E.K, who refined its content and added questions based on issues commonly encountered in routine diabetes consultations. The domains were not intended to depend equally on CGM data: some required direct interpretation of CGM metrics and trends, whereas others drew more on the broader clinical context in which CGM findings are discussed. The full question bank is provided in Multimedia Appendix 3.

For each assigned case, clinicians were asked to answer 12 questions drawn from this bank. To ensure coverage of clinically salient topics, each case included 3 questions from the blood glucose interpretation and fluctuation analysis domain, 3 questions related to the impact of food and exercise, two questions on medication and treatment guidance, two questions on emotional and psychological concerns, and one question each on long-term goals and motivation and technical issues, and device use. Questions specific to type 1 diabetes or intensive insulin therapy (for example, detailed insulin adjustment questions) were not allocated to type 2 diabetes cases. Across the 12 cases, individual question templates were reused with different vignettes and assigned to different clinicians, so that each item was answered in multiple clinical contexts while minimizing overlap in the exact question sets seen by any single clinician. Assignment details and frequency statistics for each question are provided in Multimedia Appendix 4.

In the first phase between Oct and Nov 2025, clinicians were instructed to answer their allocated questions as they would when writing to a patient in routine practice, using plain English and making reasonable assumptions based only on the CGM information and vignette provided. No formal time limit was imposed, but clinicians were asked to respond in a manner consistent with their usual clinical communication style. For each case and question, clinicians entered a single free-text response.

To promote broadly comparable levels of detail while preserving individual clinical expression, clinicians were provided with a non-mandatory guideline suggesting responses of approximately 150–250 words, with flexibility according to content complexity. No strict word limit was imposed. The CA generated one free-text response per case–question pair under identical configuration settings. No explicit word-count constraints were imposed on CA outputs; response length was determined by the model's generation process within the default API token limits. All CA outputs and clinician responses were exported and stored as separate, anonymized text units, labeled only by case ID and question ID.

In the second phase between Nov and Feb 2026, the same 6 clinicians were asked to rate anonymized responses. For each case–question pair, raters were provided with the corresponding original case materials together with an anonymized response set that included the CA answer and the clinician-authored answers written by other clinicians, so that responses were judged in the context of the underlying patient profile and CGM information rather than in isolation. Raters did not evaluate their own responses and were blinded to whether a given response had been written by a clinician or by the CA. Each response was independently rated by 3 clinicians. During the review process, clarification was also provided to all raters regarding the interpretation of certain evaluation criteria, particularly guideline adherence and actionability, when applied to more explanatory questions. Specifically, raters were advised that guideline adherence should be judged in terms of consistency with appropriate clinical practice even when no guideline was explicitly cited, and that actionability should not penalize otherwise strong explanatory responses when the question did not naturally call for specific next-step advice.

Quality was rated on 6 5-point Likert-scale dimensions (1 = very poor, 5 = excellent): clinical accuracy, guideline adherence, actionability, personalization, communication clarity, and empathy/emotional support. The operational definitions of each quality dimension and the rating scale are summarised in Table 1. Safety was captured with a 3-level flag (0 = no safety concerns, 1 = minor concern requiring revision before clinical use, 2 = major concern with unsafe or clearly contraindicated advice), and raters were also asked to guess the likely source of each response ("human clinician", "LLM", or "not sure") (Table 1). Each response, therefore, received up to 3 sets of quality scores, one safety flag, and one perceived source label for subsequent analysis.

**Table 1 Quality rating dimensions, safety flag categories, and perceived source labels.**

| Item | Definition / what raters were asked to consider | Scale/categories |
|---|---|---|
| **Clinical accuracy** | Correctness of CGM interpretation and use of numerical information (eg, TIR/TBR/TAR, mean glucose, variability indices). | 5-point Likert scale (1 = very poor, 5 = excellent) |
| **Guideline adherence** | Alignment with established diabetes and CGM practice, including consistency with major CGM and diabetes care guidelines. | 5-point Likert scale (1 = very poor, 5 = excellent) |
| **Actionability** | Clarity and feasibility of suggested next steps and contingencies for the patient, given the CGM patterns and vignette context. | 5-point Likert scale (1 = very poor, 5 = excellent) |
| **Personalization** | Explicit use of case-specific details from the vignette and CGM data, rather than generic or template-like advice. | 5-point Likert scale (1 = very poor, 5 = excellent) |
| **Communication clarity** | Ease of understanding for a layperson, including structure, wording, and avoidance of jargon or ambiguous phrasing. | 5-point Likert scale (1 = very poor, 5 = excellent) |
| **Empathy / emotional support** | Degree of validation, nonjudgmental tone, acknowledgement of | 5-point Likert scale (1 = very poor, 5 = excellent) |

| | emotional burden, and provision of supportive, encouraging language. | |
|---|---|---|
| **Safety flag** | Presence and severity of safety concerns in the advice provided (eg, unsafe insulin suggestions, failure to respond to very high or low glucose). | 0 = no safety concerns; 1 = minor concern requiring revision before clinical use; 2 = major concern with unsafe or clearly contraindicated advice |
| **Perceived source** | Rater's guess about whether the response was written by a human clinician or by the CA. | "Human clinician", "LLM", or "Not sure" |

## Statistical Analysis

Analyses were based on the clinician ratings described above. For each response, 3 raters provided 6 1–5 quality scores, a 3-level safety flag, and a perceived source label. An overall quality score was defined as the arithmetic mean of the 6 quality items for a given rating. Quality scores for CA and clinician responses were summarized using means, SDs, medians, and IQRs, overall and by question domain. Heatmaps were generated to visualize mean quality ratings across specific domains and dimensions.

Inter-rater reliability for the quality ratings was evaluated using two-way random-effects ICCs for absolute agreement [56]. Both single-rater (ICC(2,1)) and average-rater (ICC(2,3)) reliabilities were calculated separately for each quality item and for the overall quality score, with the 3 raters per response treated as interchangeable observers. 95% confidence intervals (CIs) were estimated using cluster bootstrap resampling [57] at the response level.

To compare CA and clinician performance, the 1–5 quality ratings were treated as approximately continuous, consistent with common practice for Likert-type scales in health services research. The primary analysis used linear mixed-effects models fitted separately for each quality item and for the overall quality score [58]. In these models, the individual rating was the outcome, responder type (CA vs. clinician) was included as a fixed effect with the clinician set as the reference category, and random intercepts for case and rater were included to account for the clustering of ratings within CGM cases and within individual raters. Results are reported as estimated mean differences (CA minus clinician) with 95%CIs and two-sided *P* values [59]. As a sensitivity analysis, the 3 rater scores for each response were first averaged to create matched case–question pairs. The normality of the paired differences was assessed using the Shapiro-Wilk test [60]. Because the differences deviated from a normal distribution, comparisons between CA and clinician mean scores were conducted using the non-parametric Wilcoxon signed-rank test [61].

To examine whether response length was associated with quality ratings, word count was included as a continuous fixed-effect covariate in additional linear mixed-effects models [58]. These models retained the same random-effects structure as the primary analyses to account for clustering within responses and raters. An interaction term between responder type and word count was incorporated to assess whether the association between length and quality differed between CA and clinician responses. Statistical significance of the main and interaction effects was evaluated using Wald tests [62].

Domain-specific analyses were conducted to explore whether relative performance varied by clinical topic. For each of the 6 predefined domains, the mixed-effects models were refitted on the subset of responses belonging to that domain. Because these analyses involved multiple comparisons and the study was not primarily powered for domain-level hypotheses, domain-specific results are interpreted as exploratory, with an emphasis on effect sizes and CIs rather than formal adjustment for multiplicity.

Safety flags were tabulated for CA and clinician responses and are reported descriptively, given the very low number of flagged responses. For the source-identification task (Turing test [63]), the proportions of responses that raters correctly identified were calculated both overall and at the individual rater level. Identification accuracy was compared against random chance (50%) using exact binomial tests [64]. Furthermore, to assess potential evaluation bias, overall quality scores were stratified and visualized using boxplots based on the raters' perceived source (i.e., whether the rater believed the response was generated by a CA or a clinician), regardless of the true source.

Missing ratings were rare and handled using complete case analysis without imputation. All statistical tests were two-sided (except for the one-sided binomial tests assessing accuracy greater than chance), with a significance threshold of $P < .05$. Analyses were conducted using Python 3.11.

**Reporting Framework**

This study was a simulated evaluation of an AI-based CA using de-identified real clinical CGM cases and clinician raters, rather than a randomized clinical trial. To promote transparent and reproducible reporting of the AI component, we drew partially on the SPIRIT-AI extension [65], particularly its recommendations to specify the intended use of the system, describe how clinical data are processed and presented to the model, and clarify how outputs are generated and reviewed for potential safety concerns. Elements of SPIRIT-AI that relate specifically to prospective interventional trials, such as randomization procedures, allocation concealment, trial monitoring, and adverse event reporting, were not applicable in this setting and are therefore not addressed. The checklist can be found in Multimedia Appendix 5.

# Results

## Rater Characteristics and Inter-Rater Reliability

Six senior diabetes clinicians (M:4, F:2), each with more than 10 years of post-qualification clinical experience, participated in the rating process. A total of 12 CGM-informed diabetes cases were evaluated by the 6 clinicians. Each case comprised 12 structured questions. For every case–question unit, both a CA-generated response and a clinician-authored response were assessed.

In total, 288 unique case–question responses (144 CA and 144 clinician responses) were independently evaluated. Each response was rated by exactly 3 clinicians in a blinded manner, yielding 864 response-level ratings. Each rating included 6 predefined quality dimensions (clinical accuracy, guideline adherence, actionability, personalization, communication clarity, and empathy), which were averaged to derive an overall quality score for analysis. The rating design was partially crossed but fully balanced at the response level, with no missing data. Details of clinician case assignments are provided in the Methods and Multimedia Appendix 4.

Inter-rater reliability (Table 2) was assessed using two-way random-effects ICCs for absolute agreement. For the overall quality score, single-rater reliability was fair (ICC[2,1] = 0.272; 95% CI 0.193–0.342) and increased to a moderate level when averaging 3 raters (ICC[2,3] = 0.529; 95% CI 0.418–0.609). Exploratory subgroup analyses examined inter-rater reliability separately for chatbot-authored and clinician-authored responses. For overall quality, agreement was numerically higher for chatbot-authored responses than for clinician-authored responses at both the single-rater level (ICC[2,1] 0.154 vs 0.045; ICC difference 0.109, 95% CI -0.066 to 0.159; P = .085) and the averaged-rater level (ICC[2,3] 0.354 vs 0.125; ICC difference 0.229, 95% CI -0.137 to 0.258; P=0.692), although neither difference was statistically supported.

Across individual quality dimensions, single-rater reliability was generally low to fair, ranging from 0.087 (clinical accuracy; 95% CI 0.007–0.159) to 0.324 (actionability and empathy; 95% CI 0.245–0.400 and 0.243–0.405, respectively). Reliability improved consistently when ratings were averaged across 3 clinicians, with ICC(2,3) values ranging from 0.223 (clinical accuracy; 95% CI 0.022–0.361) to 0.590 (actionability and empathy; 95% CI 0.493–0.666 and 0.491–0.671, respectively).

**Table 2 Inter-rater reliability of clinician quality ratings.**

| Quality dimension | ICC(2,1) | 95% CI | ICC(2,3) | 95% CI |
|---|---|---|---|---|
| Clinical accuracy | 0.087 | 0.007–0.159 | 0.223 | 0.022–0.361 |
| Guideline adherence | 0.102 | 0.023–0.181 | 0.254 | 0.065–0.399 |
| Actionability | 0.324 | 0.245–0.400 | 0.590 | 0.493–0.666 |
| Personalization | 0.219 | 0.141–0.290 | 0.457 | 0.330–0.551 |
| Clarity | 0.214 | 0.140–0.292 | 0.449 | 0.328–0.553 |
| Empathy | 0.324 | 0.243–0.405 | 0.590 | 0.491–0.671 |
| Overall quality score | 0.272 | 0.193–0.342 | 0.529 | 0.418–0.609 |

## Quality Comparisons

Overall quality ratings were compared between CA-generated and clinician-generated responses. Descriptive statistics were calculated using response-level scores averaged across 3 clinician raters, whereas mean differences and CIs were estimated from linear mixed-effects models. Across 288 unique case–question responses, CA responses received higher overall quality scores than clinician responses (Table 3). The mean overall quality score was 4.37 (standard deviation [SD] 0.57) for CA responses and 3.58 (SD 0.90) for clinician responses. The estimated mean difference was 0.782 (95% CI 0.692–0.872; p < 0.001). Median scores were also higher for CA responses (4.5, interquartile range [IQR] 4.0–4.8) than for clinician responses (3.8, IQR 3.0–4.2), indicating a consistent shift in central tendency. Variability differed between response types, with lower dispersion observed for CA scores (SD 0.57) relative to clinician scores (SD 0.90); this pattern suggests lower between-response dispersion in CA ratings.

**Table 3 Overall and dimension-specific quality ratings for CA and clinician responses.**

| Outcome | CA Mean (SD) | CA Median (IQR) | Clinician Mean (SD) | Clinician Median (IQR) | Mean difference (95% CI)* | P value |
|---|---|---|---|---|---|---|
| Overall quality | 4.37 (0.57) | 4.5 (4.0–4.8) | 3.58 (0.90) | 3.8 (3.0–4.2) | 0.782 (0.692–0.872) | <0.001 |
| Clinical accuracy | 4.40 (0.69) | 4.0 (4.0–5.0) | 3.84 (0.95) | 4.0 (3.0–4.0) | 0.562 (0.463–0.662) | <0.001 |
| Guideline adherence | 4.31 (0.73) | 4.0 (4.0–5.0) | 3.82 (0.92) | 4.0 (3.0–4.0) | 0.495 (0.394–0.597) | <0.001 |
| Actionability | 4.42 (0.71) | 5.0 (4.0–5.0) | 3.43 (1.13) | 4.0 (3.0–4.0) | 0.992 (0.877–1.106) | <0.001 |
| Personalization | 4.25 (0.75) | 4.0 (4.0–5.0) | 3.38 (1.11) | 4.0 (3.0–4.0) | 0.867 (0.749–0.985) | <0.001 |
| Clarity | 4.44 (0.65) | 4.0 (4.0–5.0) | 3.73 (1.04) | 4.0 (3.0–5.0) | 0.713 (0.611–0.815) | <0.001 |
| Empathy | 4.37 (0.65) | 4.0 (4.0–5.0) | 3.30 (1.19) | 3.0 (3.0–4.0) | 1.062 (0.948–1.177) | <0.001 |

*CA and clinician responses were rated on 6 quality dimensions (1–5), with the overall quality score defined as the mean of the 6 items. Values are presented as mean (SD) and median (IQR).*

*\*Mean differences (CA − clinician) were estimated using linear mixed-effects models with random intercepts for case and rater; 95% CIs and two-sided P values are shown.*

To examine whether this overall pattern was consistent across clinician-authored responses, Figure 2 presents overall quality score distributions stratified by the clinician who authored the human responses. In each stratum, the distribution of

CA scores was shifted upward relative to clinician-authored responses. CA scores restricted to the same case subsets ("matched") were comparable to the overall CA distribution, indicating that the observed difference was not confined to particular case allocations.

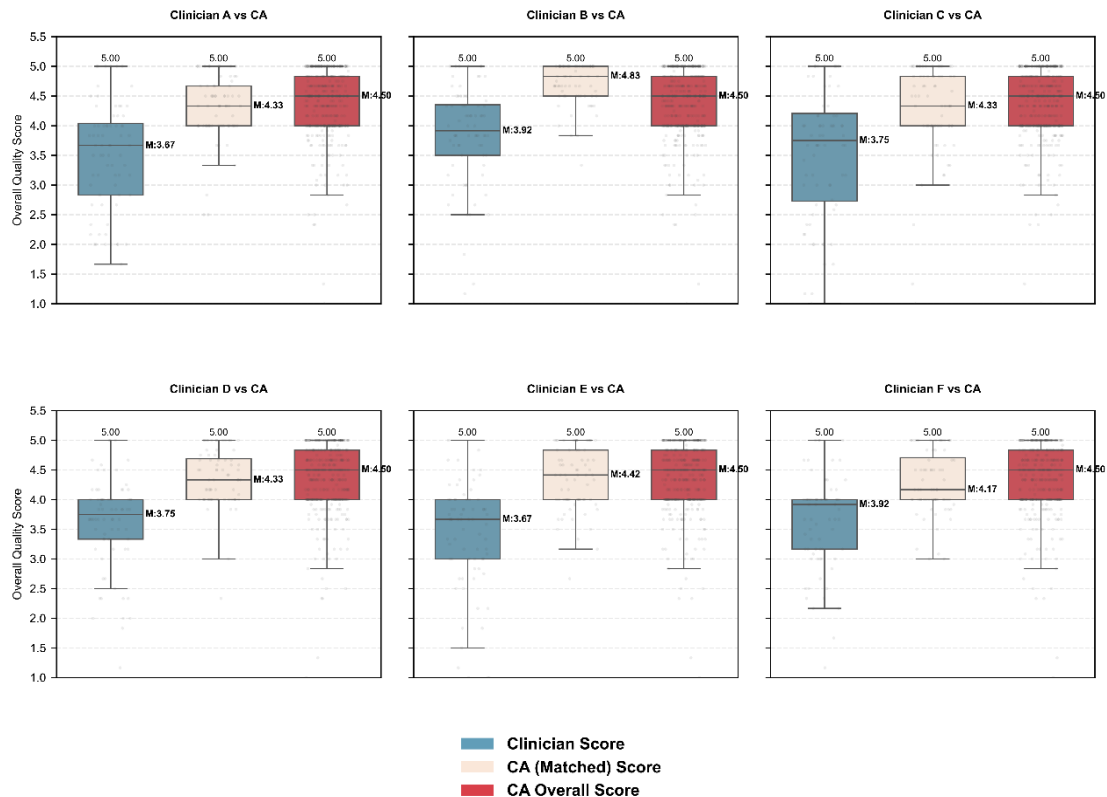

**Figure 2: Overall quality score distributions for clinician-authored and CA responses.**

*Boxplots showing response-level overall quality scores for clinician-authored and CA-generated responses, stratified by the clinician who authored the human responses (Clinicians A–F).*

*In each panel, 3 distributions are presented:*

*(i) responses written by the corresponding clinician (blue),*

*(ii) CA responses restricted to the same two cases and 24 questions assigned to that clinician ("CA matched"; beige), and*

*(iii) CA responses across all evaluated case–question responses ("CA overall"; red).*

*Overall quality scores were calculated at the response level as the mean of 6 evaluation dimensions (clinical accuracy, guideline adherence, actionability, personalization, clarity, and empathy). Each response was independently rated by 3 clinicians who did not evaluate their own authored responses.*

*Within each boxplot, the central line denotes the median (M), boxes represent the IQR, whiskers indicate the observed range, and individual points correspond to response-level scores.*

In sensitivity analyses using response-level matched case–question pairs (n = 144), the distribution of paired differences deviated from normality (Shapiro–Wilk P =

0.001). The Wilcoxon signed-rank test confirmed significantly higher scores for CA responses (W = 479, P < 0.001), with a median difference of 0.82 points. The estimated rank-biserial correlation (r ≈ 0.95) indicated a large effect size.

Response length differed markedly between CA and clinician responses. The mean word count was 211.4 (SD 54.8; 95% CI 202.4–220.4) for CA responses compared with 72.9 (SD 68.8; 95% CI 61.6–84.2) for clinician responses, representing nearly a threefold difference in verbosity.

Substantial variability in response length was also observed across individual clinicians. Mean word count ranged from 39.5 (SD 15.4; 95% CI 33.0–46.0) to 191.3 (SD 89.2; 95% CI 153.6–228.9). Inter-clinician variability was statistically significant (Kruskal–Wallis H = 65.83 [66], P < 0.001; $\varepsilon^2$ = 0.44) and remained significant after excluding the highest-verbosity clinician (H = 20.70, P < 0.001).

Despite these marked differences in response length, word count was not significantly associated with overall quality ratings in mixed-effects models (Multimedia Appendix 6). Neither the main effect of word count nor its interaction with responder type was statistically significant for the overall score (all P > 0.14). Among individual quality dimensions, only empathy demonstrated a significant interaction between word count and responder type (P = 0.003). In clinician-authored responses, word count was negatively associated with empathy ratings (β = −0.00235 per word, P < 0.001), whereas no significant association was observed for CA responses (β = 0.00107, P = 0.223). Overall, we did not find evidence that word count explained the observed differences in quality between CA and clinician responses.

Dimension-specific mixed-effects analyses identified statistically significant differences between CA and clinician responses across all 6 quality dimensions (all P < 0.001). Estimated mean differences ranged from 0.495 to 1.062 (Table 3). The largest differences were observed for empathy (1.062; 95% CI 0.948–1.177) and actionability (0.992; 95% CI 0.877–1.106), whereas smaller differences were observed for clinical accuracy (0.562; 95% CI 0.463–0.662) and guideline adherence (0.495; 95% CI 0.394–0.597). All estimated mean differences were positive.

Across the 6 predefined content domains, mixed-effects models likewise indicated statistically significant differences in overall quality between CA and clinician responses (all P ≤ 0.0248; Multimedia Appendix 7). Estimated mean differences ranged from 0.419 to 1.013. The largest estimated difference was observed in Domain A (blood glucose interpretation and fluctuation analysis; mean difference 1.013, 95% CI 0.762–1.265), followed by Domain C (medication and treatment guidance; 0.912, 95% CI 0.657–1.168).

Notably, significant differences were also observed in Domain D, which comprised emotional and psychological concerns (e.g., stress-related glycemic dysregulation, fear of hypoglycaemia, and feelings of frustration or self-doubt; estimated mean difference 0.721, 95% CI 0.507–0.935, P < 0.001). In this psychosocial domain, CA responses were consistently rated higher across quality dimensions, including empathy and actionability, indicating that performance differences were not confined to technical CGM interpretation but extended to emotionally sensitive scenarios.

The smallest estimated differences were observed in Domain E (long-term goals and motivation; 0.419, 95% CI 0.051–0.786) and Domain F (technical issues and device use; 0.493, 95% CI 0.065–0.921). Figure 3 visualizes the domain- and dimension-specific mean quality ratings, illustrating a consistent pattern of higher scores for CA responses across all 6 content domains and quality dimensions. Full descriptive statistics (mean ± SD) are provided in Multimedia Appendix 8.

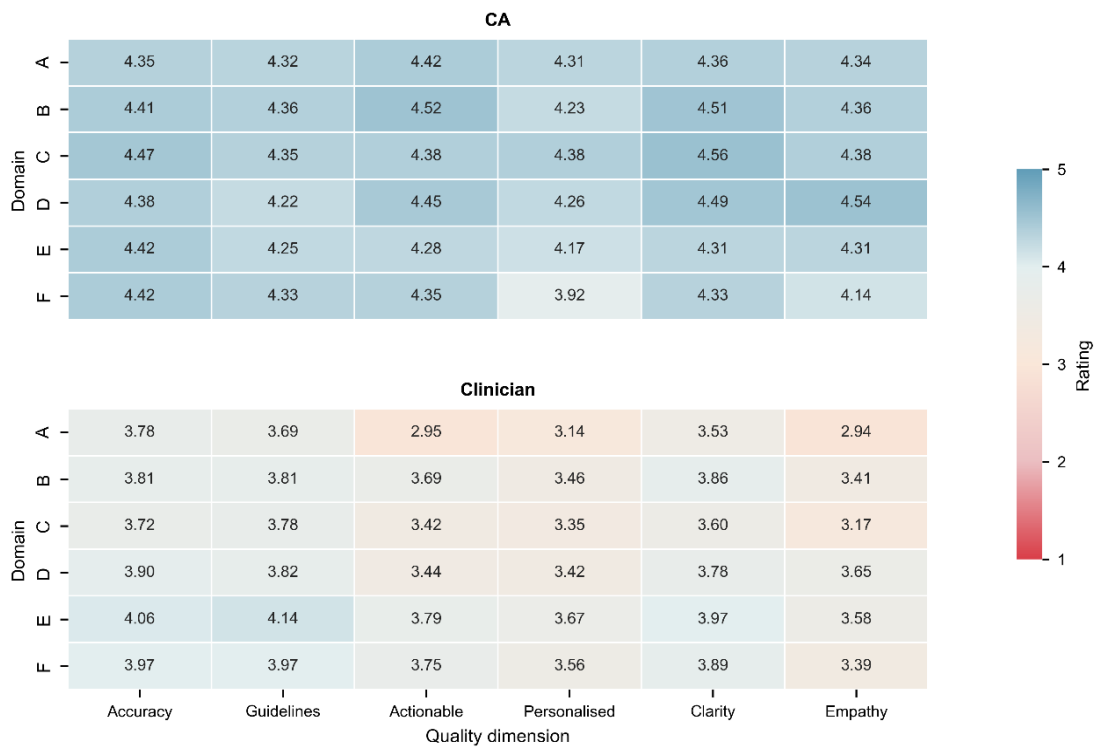

**Figure 3: Mean quality ratings by clinical domain and dimension.**

*Heatmap showing mean response-level quality ratings for CA-generated and clinician-generated responses across 6 predefined content domains (A–F) and 6 evaluation dimensions (clinical accuracy, guideline adherence, actionability, personalization, clarity, and empathy).*

*For each unique case–question response (n = 288), ratings from 3 independent clinicians were averaged to obtain a response-level score. Domain-level means were calculated by averaging these response-level scores within each domain separately for CA and clinician responses. Color intensity reflects the mean rating on a 1–5 scale.*

## Safety and source identification

Beyond quantitative quality comparisons, we examined raters' ability to identify the source of each response and the distribution of safety flags.

Across 864 ratings, clinicians correctly identified the source in 697 cases (80.7%), misclassified 100 (11.6%), and selected "not sure" in 67 (7.8%). Restricting the analysis to definitive judgments (n = 797 rating-level classifications), overall identification accuracy was 87.5% (one-sided exact binomial test versus 50% chance, P < 0.001), indicating that responses were generally distinguishable from one another.

Identification performance varied across raters. 5 clinicians demonstrated high discrimination accuracy (82.3%–100.0% among definitive judgments), whereas one clinician did not perform above chance level (50.0%; P = 0.539), suggesting substantial inter-rater heterogeneity. Rater-specific distributions of overall quality scores, stratified by perceived source (clinician vs CA), are shown in Multimedia Appendix 9.

Safety flag distributions were comparable between sources. Among clinician responses (n = 432), 387 (89.6%) were rated as Level 0, 42 (9.7%) as Level 1, and 3 (0.7%) as Level 2. Among CA responses (n = 432), corresponding proportions were 389 (90.1%), 40 (9.3%), and 3 (0.7%), respectively. However, qualitative comments were available for only 23 flagged ratings, as written explanations were optional and therefore not all safety flags were accompanied by narrative justification.

Among the 23 available comments, most (15/23, 65.2%) concerned GLP-1 eligibility under NHS body mass index (BMI) criteria, with most of these applying to CA responses (13 CA vs 2 clinicians). Three comments (13.0%) related to medication administration, specifically acarbose dosing (2 CA vs 1 clinician), and 2 comments (8.7%) noted that the response did not directly address the question. The remaining comments (n = 3, all CA) referred to behavioral feasibility concerns, including the perceived burden of the proposed action plan and the appropriateness of specific dietary substitution advice. These comment-level data should be interpreted cautiously because written remarks were optional, but they suggest that similar flag frequencies do not necessarily imply identical patterns of concern across response sources.

## Discussion

### Principal Findings

In this blinded, multi-rater evaluation of vignette-based CGM scenarios, the retrieval-grounded CA received higher mean structured quality ratings than clinician-authored responses across predefined domains. The overall mean difference was approximately 0.8 points on a 5-point scale, with the largest differences observed in empathy (mean difference 1.06) and actionability (0.99).

Under standardized vignette-based conditions, retrieval-grounded responses were judged by specialists as comparatively strong in both relational and action-oriented dimensions. To our knowledge, this represents one of the first blinded comparative evaluations of retrieval-grounded LLM-generated and clinician-authored responses in structured CGM counseling contexts.

Performance differences were not uniform across domains and appeared to vary according to task structure. Differences were most pronounced in data-intensive glucose interpretation and action-oriented explanation, and attenuated in domains centred on long-term motivational support and device troubleshooting. Notably, significant differences were also observed in psychosocial scenarios, indicating that comparative ratings were not limited to quantitative glucose analysis but extended to contexts involving emotional distress, fear of hypoglycaemia and diabetes-related frustration. This gradient suggests that relative performance may depend on the cognitive structure of the task. Retrieval-grounded CAs may be particularly well suited to synthesizing numerical CGM metrics and guideline-based reasoning into structured explanations and action plans [25]. In contrast, domains requiring nuanced behavioral coaching or experiential clinical judgment may depend more heavily on individualized framing [67,68].

**Comparison With Prior Work**

These findings extend prior CGM-focused evaluations of LLM-generated summaries, which have typically assessed model outputs in isolation and emphasised feasibility, accuracy, or clinical acceptability [31]. By employing a balanced design with blinded ratings, the present study enables direct comparison with clinician-authored responses across predefined technical and relational dimensions. Notably, the largest differences were observed in actionability and empathy, rather than being confined solely to glycemic interpretation, suggesting that structured explanatory and relational components may be particularly sensitive to comparative evaluation.

Related work in other wearable-data domains has explored the use of AI systems to interpret structured physiological data streams, such as heart rate, physical activity, or sleep measures generated by consumer wearables [69-71]. These studies have similarly focused on automated summarization, health insight generation, or behavioral coaching based on wearable-derived data [70]. However, most have evaluated system outputs in isolation or against guideline-based expectations rather than through blinded comparison with clinician-authored responses. The present study therefore contributes additional evidence by examining how LLM-generated explanations compare directly with clinician communication in a structured evaluation setting.

The evaluation was confined to structured, vignette-based scenarios involving common CGM-related questions, concept clarification, and general diabetes self-management guidance. It did not assess complex therapeutic decision-making, individualized prescribing adjustments, or real-time risk management. Accordingly, these findings should be interpreted within the scope of clearly bounded explanatory tasks, rather than extended to higher-stakes clinical reasoning contexts.

Beyond CGM-specific evaluations, these findings align with broader diabetes care literature, suggesting cautious optimism regarding AI-supported communication tools. Reported areas of potential utility include patient education, explanation of structured health data and information synthesis, whereas greater caution is typically expressed when systems are used to support individualized treatment decisions or medication adjustments [68,72]. Within this context, retrieval-grounded CAs may be best viewed as adjunct tools for patient-facing explanation in clearly defined, guideline-constrained use cases, with clinician oversight maintained for interpretation and final clinical framing [31,67-72].

Although workflow outcomes were not directly evaluated, the observed performance in standardized CGM explanation tasks suggests potential relevance to clinical workflow. In routine practice, summarising CGM trends, clarifying commonly used thresholds and addressing frequent educational queries can consume substantial consultation time [46]. If deployed within appropriately bounded tasks and aligned to local guidance and policy constraints, retrieval-grounded CAs may help support these informational components of care. Any such use should be framed as supportive rather than substitutive, and prospective studies are needed to quantify effects on clinician workload, consultation flow and patient outcomes.

**Limitations**

Several factors limit how these findings should be interpreted and generalized. Inter-rater agreement was modest, particularly for clinical accuracy, reflecting variability in expert judgments of complex narrative clinical responses, although reliability improved when ratings were averaged across 3 clinicians. At the aggregate level, safety flag frequencies were similar between CA and clinician responses, suggesting that the overall proportion of flagged responses was comparable under the study rubric. However, most optional written comments were attached to CA responses (20/23, 87%). This asymmetry should be interpreted cautiously. Safety judgments in CGM counseling depend not only on factual correctness, but also on local prescribing criteria, practice norms, and individual clinicians' thresholds for caution [73] (eg, GLP-1 eligibility under NHS BMI criteria [74]). In a multi-rater setting that may include clinicians trained in different systems, some disagreement is therefore expected [75]. Nonetheless, the concentration of comments on CA outputs suggests that similar flag frequencies should not be interpreted as evidence of identical safety profiles across response sources. Clinicians more often documented specific concerns for CA responses, including policy nonalignment, incomplete question addressing, and the feasibility or appropriateness of suggested actions. Although these issues were infrequent, they remain clinically important because even occasional mismatches may introduce safety risk in patient-facing use if not identified.

While adjusted analyses did not indicate that response length independently explained the overall score differences, the greater length and stylistic characteristics of CA responses may still have shaped clinicians' perceptions, particularly of relational quality. The interaction between responder type and word count for empathy is consistent with this possibility. Moreover, the high rate of correct source identification suggests that CA and clinician responses were often distinguishable on structural or stylistic grounds. Residual perceptual bias, therefore, cannot be excluded.

Results reflect a single retrieval-grounded system and one model configuration (GPT-5.1), and may differ under alternative models, retrieval resources or prompting strategies. Although vignettes were derived from real CGM data and reviewed by clinicians, they necessarily simplify real-world encounters and did not encompass rare diseases, atypical presentations or highly complex comorbid scenarios. The case set focused primarily on common, guideline-constrained CGM interpretation and self-management questions, which may favor structured retrieval-based synthesis. Comparative performance may therefore differ in diagnostically ambiguous or therapeutically complex contexts. Furthermore, written vignettes capture communication quality under controlled conditions rather than interactive consultations. Although raters were instructed to score how well responses addressed the question as asked, consistent application of this guidance cannot be independently verified.

Future work should extend this work beyond vignette-based evaluation into prospective real-world clinical settings. In particular, it would be valuable to assess the system as part of routine diabetes clinics, for example in preconsultation preparation or patient education workflows, where larger and more diverse samples could be recruited and where outcomes such as patient understanding, question generation, consultation efficiency, clinician workload, and perceived usefulness could be evaluated more directly. Such studies would also allow assessment of how patients engage with different response formats, including concise bullet point summaries versus more detailed explanatory prose.

At the same time, future research should address the governance requirements of real-world deployment. This includes evaluating the system under local clinical policies, defining appropriate oversight and escalation pathways, and examining safety in interactive use rather than single-turn written responses alone. Because implementation in live clinics would raise important ethical, regulatory, and data governance issues, a phased approach may be most appropriate, beginning with supervised observational or pilot studies before progressing to broader prospective evaluation.

**Conclusions**

Taken together, these findings support a potential role for retrieval-grounded LLM systems as adjunct tools for structured CGM interpretation and patient-facing explanation of common questions arising in routine diabetes care under controlled vignette-based conditions. In practical terms, their most appropriate role appears to be explanatory and educational rather than therapeutic: they may help patients understand glucose patterns, clarify common concerns, and prepare for discussion with their diabetes care team, including in relation to emotionally sensitive issues. However, this should not be interpreted as support for individualized therapeutic decision-making. In particular, such systems are not established here as appropriate for recommending medication initiation, dose adjustment, regimen change, or other personalized treatment decisions, which should remain clinician-led. Nor do these findings establish equivalence to clinician communication in real-world consultations, where ongoing therapeutic relationships, contextual judgment, and dynamic interaction remain central. Further prospective evaluation in interactive clinical settings is needed to define the appropriate scope, safeguards, and implementation of such systems in practice.

# Acknowledgements

The authors would like to thank all participants who contributed their time and effort to this study. We also sincerely thank the 6 senior diabetes clinicians for their valuable support, expert input, and careful contribution to this work. Their clinical expertise, thoughtful feedback, and engagement in the evaluation process were essential to the development and refinement of this study. We are also grateful to colleagues and collaborators who provided helpful discussions and support during the design, implementation, and preparation of this research.

# Funding Statement

This study was supported by internal project funding from the UCL Institute of Health Informatics, the University College London Hospitals Biomedical Research Centre, and the Moorfields Biomedical Research Centre. The funders had no role in the study design; data collection, analysis, or interpretation; manuscript writing; or the decision to submit the article for publication.

## Conflicts of Interest

The authors declare no financial competing interests. Several coauthors were involved in the study as blinded raters and/or contributors of comparator responses. To minimize potential bias, raters were blinded to response source and did not assess their own responses. Their involvement in manuscript review and revision was limited to interpretation, clinical contextualization, and critical review of the manuscript.

## Data Availability

The ShanghaiT1DM and ShanghaiT2DM datasets are publicly available via Figshare (https://figshare.com/articles/dataset/Diabetes_Datasets-ShanghaiT1DM_and_ShanghaiT2DM/20444397). The OhioT1DM dataset is publicly available via Kaggle (https://www.kaggle.com/datasets/ryanmouton/ohiot1dm).

The curated retrieval corpus used for the RAG system, derived exclusively from publicly available clinical guidelines and educational materials, is available at: https://github.com/candiceguo0528/Diabetes-chatbot. No new patient data were generated in this study.

## Code availability

The CA implementation, including the RAG pipeline and analysis scripts used in this study, is available at: https://github.com/candiceguo0528/Diabetes-chatbot.

## Author Contributions

**Contributor**

ZG and KL contributed to the conception and design of the study. ZG led the development and deployment of the system. ZG and KL coordinated participant recruitment and study facilitation, with support from clinical collaborators through professional referral and study dissemination. EK reviewed the question bank, assessed the patient profiles, and provided targeted clinical feedback during early-stage system development. EK, AV, IA, CA, and JH provided clinical input and expert feedback on the study design, system development, and manuscript. ZG wrote the first draft of the manuscript. All authors contributed to the review and revision of the manuscript. All authors read and approved the final version of the manuscript.

**Corresponding author**

Correspondence to Kezhi Li.

# Abbreviations

**ADA**: American Diabetes Association
**AGP**: ambulatory glucose profile
**API**: application programming interface
**BMI**: body mass index
**CA**: conversational agent
**CGM**: continuous glucose monitoring
**CI**: confidence interval
**FAISS**: Facebook AI Similarity Search
**GMI**: glucose management indicator
**ICC**: intraclass correlation coefficient
**IDF**: International Diabetes Federation
**IQR**: interquartile range
**LLM**: large language model
**NHS**: National Health Service
**RAG**: retrieval-augmented generation
**SD**: standard deviation
**TAR**: time above range
**TBR**: time below range
**TIR**: time in range
**UCLH**: University College London Hospitals

# Multimedia Appendix 1

Patient Name: Steven (ID: 1002)
Date: Based on CGM Data – May 2021

1. Demographics & Medical History

- Age: 36 years
- Sex: Male
- Height / Weight / BMI: 172 cm / 68 kg / BMI: 23.0
- Occupation: Graphic designer (mostly sedentary, 8+ hours/day screen time)
- Living Situation: Lives with partner; meals are prepared at home on weekdays, dine-out on weekends
- Diabetes Type: Type 1 Diabetes Mellitus (diagnosed at age 30)
- Duration: 6 years since diagnosis
- Family History: Father has hypertension; no known family history of type 1 or type 2 diabetes
- Diabetes Education: Attended structured DSME course in 2020; currently self-managing pump and CGM independently

2. Treatment & Medication

- Insulin Regimen:
    - Delivery: Continuous Subcutaneous Insulin Infusion (CSII) – Novolin R
    - Basal Insulin:
        - Mean basal rate: 0.68 IU/h
        - Median basal rate: 0.7 IU/h
    - Bolus Insulin:
        - Total bolus injections recorded: 28
        - Mean bolus dose: 5.46 IU
        - Administered manually before meals based on experience (no auto calculator)
- Other Antihyperglycemics: None
- Adherence:
    - Wears CGM regularly (>89% wear time in recent monitoring)
    - Consistently inputs bolus doses; manages basal profiles without assistance
    - Does not currently count carbohydrates or use an insulin-to-carb ratio (ICR)

3. CGM Monitoring Summary

Monitoring duration: 11 days
CGM wear time: 89.8%
Total glucose readings: 948

| Metric | Value | Recommended Target* |
|---|---|---|
| Mean glucose | 136.0 mg/dL (7.6 mmol/L) | — |
| GMI (Estimated HbA1c) | 6.56% | — |
| Standard deviation | 59.9 mg/dL (3.3 mmol/L) | — |
| Coefficient of variation | 44.0% | <36% |
| Time in Range (70–180 mg/dL [3.9–10.0 mmol/L]) | 74.8% | >70% |
| Time Below Range (<70 mg/dL [<3.9 mmol/L]) | 7.1% | <4% |
| Time in Very Low (<54 mg/dL [<3.0 mmol/L]) | 0.8% | <1% |
| Time Above Range (>180 mg/dL [>10.0 mmol/L]) | 18.1% | <25% |
| Time in Very High (>250 mg/dL [>13.9 mmol/L]) | 5.9% | <5% |

* Based on international CGM consensus targets

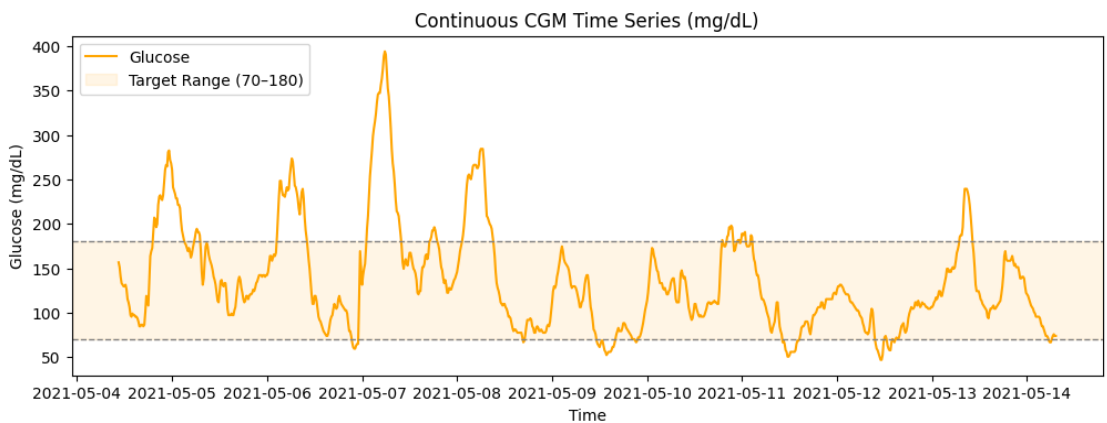

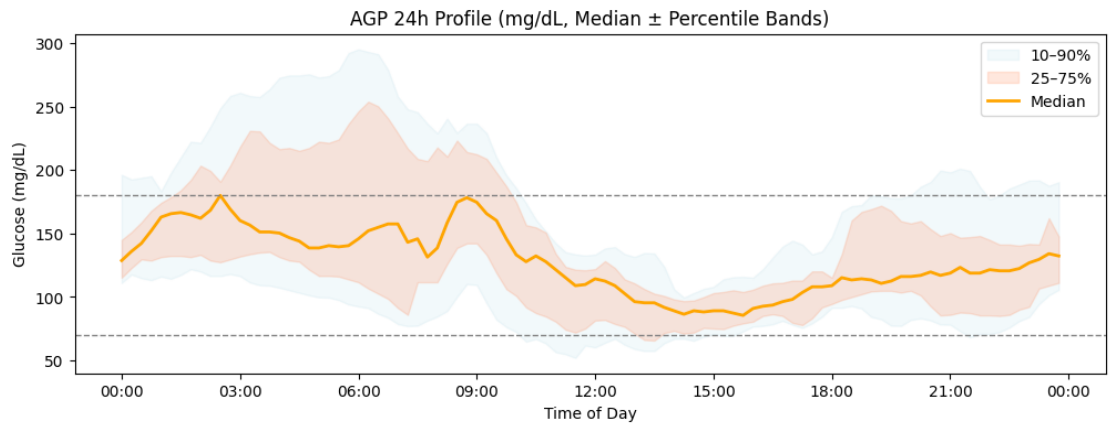

4. Lifestyle & Daily Routine

- Diet:
    - Weekdays: three regular meals at home
        - Breakfast (8:30 AM): oatmeal with milk, egg or toast
        - Lunch (1:00 PM): rice/noodles with vegetables and meat
        - Dinner (7:00 PM): usually stir-fried dishes with rice or soup
    - Weekends: dine out (hotpot, fast food, or noodles), heavier carbohydrate load
    - Occasional afternoon snack (~4 PM): fruit, yogurt, or biscuit
    - Avoids sugary drinks except for hypoglycemia treatment (uses juice or glucose tabs)
    - Alcohol: ~1–2 drinks per month, typically with meals
    - Does not smoke
- Physical Activity:
    - Walks 15–20 minutes daily (commute and errands)
    - No structured exercise routine
    - Does not pre-adjust insulin for exercise; monitors symptoms and eats snacks as needed
- Sleep:
    - Sleep from ~12:30 AM to 7:30 AM on weekdays
    - Slightly longer on weekends
    - Occasionally reports post-lunch fatigue
- Stress & Work:
    - Moderate work stress due to deadlines
    - Seated >8 hours daily due to nature of desk job

5. Self-Management Behavior

- Insulin Administration:
    - Comfortable using CSII independently
    - Adjusts basal settings when necessary (e.g., illness, heavy meals)
    - Does not use bolus calculator; estimates boluses based on food size/type
    - Fingerstick glucose checks: 2–3 times/week, especially to confirm CGM readings <70 mg/dL (<3.9 mmol/L) and to rule out falsely low readings
- Monitoring Practices:
    - Uses FreeStyle Libre daily; scans CGM at least every 8 hours
    - Uploads data to LibreView before appointments
    - Occasionally performs fingerstick blood glucose (2–3 times/week), especially when feeling low
- Record Keeping:
    - No written food logs
    - No use of mobile apps for dose tracking or meal logs
    - No documented use of ICR/ISF ratios or correction bolus formulas
- Education Status:
    - Completed diabetes self-management education once
    - No recent refresher but open to follow-up sessions
    - Has not seen a dietitian in past 2 years

6. Reported Challenges & Self-Observations

- Notices afternoon lows (~3–4 PM), especially on days with walking or delayed lunch
- Experiences occasional spikes after weekend meals
- Sometimes fatigued in late morning or post-lunch
- No severe hypoglycemia requiring assistance
- Finds carb counting tedious; prefers consistent meals to simplify insulin dosing
- Would like to improve post-meal stability, especially after lunch and weekend dinners

7. Clinical Readiness & Patient Engagement

The patient demonstrates high engagement with diabetes technology and self-care routines. He uses a CGM regularly with >89% wear time and inputs insulin doses consistently via CSII. He monitors glucose fluctuations and is comfortable adjusting basal rates manually.

Although he does not currently practice carbohydrate counting or use an ICR/ISF model, he maintains stable daily routines and reports meal-based decision-making. He performs occasional capillary glucose checks in specific situations (e.g., symptoms, driving) and uploads CGM data proactively.

He does not maintain written logs of meals or insulin, but his meal patterns are regular. He is receptive to additional education or nutrition counseling and is interested in improving day-to-day glucose stability without complicating his routine. No behavioral red flags or self-care barriers are currently reported.

## Multimedia Appendix 2

INITIAL_ANALYSIS_PROMPT = """You are GluMentor, a board-certified diabetes specialist.

You are reviewing a Continuous Glucose Monitoring (CGM) report in the context of the patient profile provided below.

The patient may be feeling anxious, uncertain, or overwhelmed, so your job is to explain the results clearly, warmly, and responsibly.

PATIENT PROFILE
─────────────────────

{patient_info}

CGM DATA SUMMARY
─────────────────────

{summary}

Please provide a supportive and clinically grounded interpretation in the style of a real doctor speaking to a patient during a calm, face-to-face consultation.

Write in complete, natural paragraphs rather than bullets. Use plain, accessible language and avoid unnecessary jargon.

Your response should:

begin with a warm acknowledgement of the patient's effort and a brief overall impression of what the glucose data shows;

explain the average glucose and estimated HbA1c (GMI) in simple terms, including what they may suggest about longer-term glucose exposure;

explain variability, including SD and CV, in plain language, focusing on what "stable" or "swinging" glucose means in daily life;

interpret Time in Range and explain what {tir:.1f}% means in a way a patient can easily understand;

highlight any important high or low glucose patterns, including likely timing if supported by the data;

gently identify the most important priority areas for improvement without sounding critical or alarming;

offer 3 to 5 practical next steps that are clearly tailored to the patient profile and CGM findings.

Finish by gently inviting the patient to share what they would like to focus on next in their diabetes journey.

Base your interpretation only on the information provided. Do not invent symptoms, medications, diagnoses, or lifestyle details that are not stated.

Do not prescribe medication changes or make definitive diagnoses from CGM data alone.

If the data suggests potentially concerning patterns, acknowledge them calmly and encourage appropriate follow-up with the diabetes care team."""

FOLLOWUP_PROMPT = """You are GluMentor, continuing a diabetes consultation.

PATIENT PROFILE
─────────────────────

{patient_info}

CGM DATA SUMMARY
─────────────────────

{summary}

USER QUESTION
─────────────────────

{user_input}

REFERENCE INFORMATION
─────────────────────

{rag_context}

Please answer the user as a supportive, thoughtful diabetes specialist speaking in a natural follow-up conversation.

Respond directly to the user's concern without repeating the full CGM summary. You may briefly refer to relevant CGM findings if they help answer the question clearly.

Use warm, calm, human language that builds trust and emotional safety. Write in complete, flowing paragraphs, not bullet points.

Your response should:

address the user's actual question first, rather than giving a generic review;

connect your advice to the patient profile, CGM patterns, and any relevant reference information when available;

provide specific, practical, and realistic guidance rather than vague suggestions;

explain reasoning in plain language when useful, especially for food choices, timing, patterns, and self-management decisions;

maintain a professional and compassionate tone, as if you are a caring doctor sitting across from the patient.

Base your answer only on the information provided. Do not invent details that are not stated.

Do not make medication changes, insulin dosing recommendations, or definitive diagnoses unless such information is explicitly supported and the system is designed for that purpose.

If the question raises safety concerns or suggests urgent symptoms, advise timely contact with an appropriate healthcare professional in a calm and responsible way."""

# Multimedia Appendix 3

**A: Blood Glucose Interpretation & Fluctuation Analysis**

1. "Are my glucose readings normal? Why do they fluctuate like this?"
2. "Why does my blood sugar spike after breakfast?"
3. "Is my TIR good enough?"
4. "Why do I drop low during the night even when I sleep normally?"
5. "My blood sugar is all over the place this week, what's going on?"
6. ''Whatever I do, I cannot bring down my fasting blood sugar.''

**B: Impact of Food & Exercise**

1. "I ate or exercised this way, but why did my blood sugar change like that?"
2. "I had noodles and my sugar shot up to 13, should I avoid carbs completely?"
3. "Should I exercise before or after meals?"
4. "Are low-calorie or zero-sugar drinks actually safe?"
5. ''I cannot remember to bolus before meals, I do not always know if I am going to eat within the next 20 minutes.''
6. ''I often have hypos after meals, why does this happen?''
7. '' What to do with alcohol?''

**C: Medication & Treatment Guidance**

1. "Is my medication working? Do I need to change anything?
2. "I'm taking meds, but my sugar is still high, why?"
3. "Am I going to need insulin soon?"
4. "Could it be that I'm taking my pills too early or late?"

**D: Emotional & Psychological Concerns**

1. "My stress or mood affects my sugar, what should I do?"
2. "I've been under a lot of stress and my blood sugar is a mess."
3. "I've worked so hard but my glucose is still bad, I feel defeated."
4. "Other people are doing better than me. Am I failing?"
5. ''I am very afraid of hypos and I do less insulin than I should.''

**E: Long-term Goals & Motivation**

1. "Can I really get better? What should I aim for?"
2. "I want my HbA1c below 6.5%, how can I achieve it?"
3. "Is there any chance I can get off medications?"
4. "Once I start insulin, is it forever?"
5. ''I definitely want to avoid complications, but I find it hard to achieve 100% TIR.''

**F: Technical Issues & Device Use**

1. "Is my CGM device working properly?"
2. "Is this spike real, or is the sensor wrong?"
3. "My sensor keeps falling off, what can I do?"
4. "Why is there a gap in my readings last night?"

# Multimedia Appendix 4

**S 1: Clinician case assignments, question-ID coverage, and reviewed cases**

| Clinician | Completed Cases | Question IDs Covered in Completed Cases | Cases Reviewed (6 per clinician) |
|---|---|---|---|
| D1 | Case 1, Case 7 | **Case 1:** A1, A2, A3, B1, B3, B5, C1, C2*, D1, D5, E1, F1 **Case 7:** A4, A5, A6, B2, B4, B6*, C3, C4, D2, D3, E2, F3 | Case 2, Case 3, Case 4, Case 5, Case 6, Case 8 |
| D2 | Case 2, Case 8 | **Case 2:** A1, A5, A6, B5, B6, B7, C1, C4, D1, D4, E3, F4 **Case 8:** A2, A4, A5, B2, B3, B4, C2, C3, D2, D3, E2, F2 | Case 1, Case 3, Case 4, Case 7, Case 10, Case 12 |
| D3 | Case 3, Case 9 | **Case 3:** A3, A4, A5, B3, B4, B5, C1, C4, D3, D5, E4, F3 **Case 9:** A1, A2, A6, B1, B6, B7, C3, C2, D2, D4, E3, F1 | Case 1, Case 2, Case 6, Case 7, Case 8, Case 11 |
| D4 | Case 4, Case 10 | **Case 4:** A4, A5, A6, B1, B2, B7, C1, C4, D4, D5, E4, F4 **Case 10:** A1, A2, A3, B1, B6, B7, C2, C3, D1, D2, E5, F2 | Case 1, Case 5, Case 6, Case 9, Case 11, Case 12 |
| D5 | Case 5, Case 11 | **Case 5:** A1, A5, A6, B1, B2, B3, C1, C4, D1, D3, E5, F1 **Case 11:** A3, A4, A5, B2, B4, B6, C2, C3, D2, D4, E1, F3 | Case 2, Case 3, Case 7, Case 9 Case 10, Case 12 |
| D6 | Case 6, Case 12 | **Case 6:** A1, A2, A6, B1, B4, B5, C2, C4, D1, D3, E1, F2 **Case 12:** A3, A4, A5, B3, B6, B7, C1, C3, D3, D4, E2, F4 | Case 4, Case 5, Case 8, Case 9, Case 10, Case 11 |

**S 2: Frequency of each question ID across all 12 cases (N = 144)**

| Question ID | Count | Question ID | Count |
|---|---|---|---|
| A1 | 6 | C1 | 6 |
| A2 | 5 | C2 | 6 |
| A3 | 5 | C3 | 6 |
| A4 | 6 | C4 | 6 |
| A5 | 8 | D1 | 5 |
| A6 | 6 | D2 | 5 |
| B1 | 6 | D3 | 6 |
| B2 | 5 | D4 | 5 |
| B3 | 5 | D5 | 3 |
| B4 | 5 | E1 | 3 |
| B5 | 4 | E2 | 3 |
| B6 | 6 | E3 | 2 |
| B7 | 5 | E4 | 2 |
|  |  | E5 | 2 |
| F1 | 3 | F3 | 3 |

| **F2** | 3 | **F4** | 3 |

# Multimedia Appendix 5

## SPIRIT 2025 checklist of items to address in a randomized trial protocol

| Section / Topic | No | SPIRIT 2025 checklist item description | Reported on page no. |
|---|---|---|---|
| **Administrative information** | | | |
| Title and structured summary | 1a | Title stating the trial design, population, and interventions, with identification as a protocol | 1 |
| | 1b | Structured summary of trial design and methods, including items from the World Health Organization Trial Registration Data Set | 1–2 |
| Protocol version | 2 | Version date and identifier | |
| Roles and responsibilities | 3a | Names, affiliations, and roles of protocol contributors | 1; 26 |
| | 3b | Name and contact information for the trial sponsor | 1 |
| | 3c | Role of trial sponsor and funders in design, conduct, analysis, and reporting of trial; including any authority over these activities | 25 |
| | 3d | Composition, roles, and responsibilities of the coordinating site, steering committee, endpoint adjudication committee, data management team, and other individuals or groups overseeing the trial, if applicable | 26 |
| **Open science** | | | |
| Trial registration | 4 | Name of trial registry, identifying number (with URL), and date of registration. If not yet registered, name of intended registry | |
| Protocol and statistical analysis plan | 5 | Where the trial protocol and statistical analysis plan can be accessed | 26 |
| Data sharing | 6 | Where and how the individual de-identified participant data (including data dictionary), statistical code, and any other materials will be accessible | 26 |
| Funding and conflicts of interest | 7a | Sources of funding and other support (e.g., supply of drugs) | 25 |
| | 7b | Financial and other conflicts of interest for principal investigators and steering committee members | 25 |
| Dissemination policy | 8 | Plans to communicate trial results to participants, healthcare professionals, the public, and other relevant groups (e.g., reporting in trial registry, plain language summary, publication) | |

| **Introduction** | | | |
|---|---|---|---|
| Background and rationale | 9a | Scientific background and rationale, including summary of relevant studies (published and unpublished) examining benefits and harms for each intervention | 3–5 |
| | 9b | Explanation for choice of comparator | 5-6 |
| Objectives | 10 | Specific objectives related to benefits and harms | 6 |
| **Methods: Patient and public involvement, trial design** | | | |
| Patient and public involvement | 11 | Details of, or plans for, patient or public involvement in the design, conduct, and reporting of the trial | 11-13 |
| Trial design | 12 | Description of trial design including type of trial (e.g., parallel group, crossover), allocation ratio, and framework (e.g., superiority, equivalence, non-inferiority, exploratory) | 11-13 |
| **Methods: Participants, interventions, and outcomes** | | | |
| Trial setting | 13 | Settings (e.g., community, hospital) and locations (e.g., countries, sites) where the trial will be conducted | 11 |
| Eligibility criteria | 14a | Eligibility criteria for participants | 11 |
| | 14b | If applicable, eligibility criteria for sites and for individuals who will deliver the interventions (e.g., surgeons, physiotherapists) | 11 |
| Intervention and comparator | 15a | Intervention and comparator with sufficient details to allow replication including how, when, and by whom they will be administered. If relevant, where additional materials describing the intervention and comparator (e.g., intervention manual) can be accessed | 8-13 |
| | 15b | Criteria for discontinuing or modifying allocated intervention/comparator for a trial participant (e.g., drug dose change in response to harms, participant request, or improving/worsening disease) | |
| | 15c | Strategies to improve adherence to intervention/comparator protocols, if applicable, and any procedures for monitoring adherence (e.g., drug tablet return, sessions attended) | |
| | 15d | Concomitant care that is permitted or prohibited during the trial | |
| Outcomes | 16 | Primary and secondary outcomes, including the specific measurement variable (e.g., systolic blood pressure), analysis metric (e.g., change from baseline, final value, time to event), method of aggregation | 12-13 |

| | | (e.g., median, proportion), and time point for each outcome | |
|---|---|---|---|
| Harms | 17 | How harms are defined and will be assessed (e.g., systematically, non-systematically) | |
| Participant timeline | 18 | Time schedule of enrollment, interventions (including any run-ins and washouts), assessments, and visits for participants. A schematic diagram is highly recommended (see Figure) | |
| Sample size | 19 | How sample size was determined, including all assumptions supporting the sample size calculation | 12-15 |
| Recruitment | 20 | Strategies for achieving adequate participant enrollment to reach target sample size | |
| **Methods: Assignment of interventions** | | | |
| Randomization: | | | |
| Sequence generation | 21a | Who will generate the random allocation sequence and the method used | 11-14; 43 |
| | 21b | Type of randomization (simple or restricted) and details of any factors for stratification. To reduce predictability of a random sequence, other details of any planned restriction (e.g., blocking) should be provided in a separate document that is unavailable to those who enroll participants or assign interventions | |
| Allocation concealment mechanism | 22 | Mechanism used to implement the random allocation sequence (e.g., central computer/telephone; sequentially numbered, opaque, sealed containers), describing any steps to conceal the sequence until interventions are assigned | 11-14 |
| Implementation | 23 | Whether the personnel who will enroll and those who will assign participants to the interventions will have access to the random allocation sequence | |
| Blinding | 24a | Who will be blinded after assignment to interventions (e.g., participants, care providers, outcome assessors, data analysts) | 11-14 |
| | 24b | If blinded, how blinding will be achieved and description of the similarity of interventions | 11-14 |
| | 24c | If blinded, circumstances under which unblinding is permissible, and procedure for revealing a participant's allocated intervention during the trial | 11-14 |
| **Methods: Data collection, management, and analysis** | | | |
| Data collection methods | 25a | Plans for assessment and collection of trial data, including any related processes to promote data quality (e.g., duplicate measurements, training of assessors) and a description of trial instruments (e.g., questionnaires, laboratory tests) along with their reliability and validity, if known. Reference to where data collection forms can be accessed, if not in the protocol | 11-15 |
| | 25b | Plans to promote participant retention and complete follow-up, including list of any outcome data to be collected for participants who discontinue or deviate from intervention protocols | |
| Data management | 26 | Plans for data entry, coding, security, and storage, including any related processes to promote data quality (e.g., double data entry; range checks for data values). Reference to where details of data management procedures can be accessed, if not in the protocol | |
| Statistical methods | 27a | Statistical methods used to compare groups for primary and secondary outcomes, including harms | 14-15 |
| | 27b | Definition of who will be included in each analysis (e.g., all randomized participants), and in which group | 14-15 |
| | 27c | How missing data will be handled in the analysis | 14-15 |
| | 27d | Methods for any additional analyses (e.g., subgroup and sensitivity analyses) | 14-15 |
| **Methods: Monitoring** | | | |
| Data monitoring committee | 28a | Composition of data monitoring committee (DMC); summary of its role and reporting structure; statement of whether it is independent from the sponsor and funder; conflicts of interest and reference to where further details about its charter can be found, if not in the protocol. Alternatively, an explanation of why a DMC is not needed | |
| | 28b | Explanation of any interim analyses and stopping guidelines, including who will have access to these interim results and make the final decision to terminate the trial | |
| Trial monitoring | 29 | Frequency and procedures for monitoring trial conduct. If there is no monitoring, give explanation | 11-13 |
| **Ethics** | | | |
| Research ethics approval | 30 | Plans for seeking research ethics committee/institutional review board approval | 6 |
| Protocol amendments | 31 | Plans for communicating important protocol modifications to relevant parties | |
| Consent or assent | 32a | Who will obtain informed consent or assent from potential trial participants or authorized proxies, and how | |
| | 32b | Additional consent provisions for collection and use of participant data and biological specimens in ancillary studies, if applicable | |
| Confidentiality | 33 | How personal information about potential and enrolled participants will be collected, shared, and maintained in order to protect confidentiality before, during, and after the trial | 6; 10 |
| Ancillary and post-trial care | 34 | Provisions, if any, for ancillary and post-trial care, and for compensation to those who suffer harm from trial participation | |

## Multimedia Appendix 6

**S 3: Association between response length and quality ratings**

| Outcome | CA β (95% CI) | P value | Clinician β (95% CI) | P value | Interaction P value |
|---|---|---|---|---|---|
| Overall quality | 0.00072 (−0.0006 to 0.0021) | 0.293 | −0.00055 (−0.0016 to 0.0005) | 0.305 | 0.149 |
| Clinical accuracy | 0.00013 (−0.0013 to 0.0016) | 0.858 | 0.00002 (−0.0011 to 0.0012) | 0.979 | 0.902 |
| Guideline adherence | 0.00125 (−0.0002 to 0.0027) | 0.094 | −0.00052 (−0.0017 to 0.0006) | 0.379 | 0.068 |
| Actionability | 0.00027 (−0.0014 to 0.0019) | 0.753 | −0.00038 (−0.0017 to 0.0009) | 0.574 | 0.557 |
| Personalization | 0.00081 (−0.0009 to 0.0025) | 0.345 | −0.00055 (−0.0019 to 0.0008) | 0.413 | 0.211 |
| Clarity | 0.00041 (−0.0012 to 0.0020) | 0.623 | 0.00053 (−0.0007 to 0.0018) | 0.407 | 0.912 |
| Empathy | 0.00107 (−0.0006 to 0.0028) | 0.223 | −0.00235 (−0.0037 to −0.0010) | 0.001 | 0.003 |

Linear mixed-effects models were fitted separately for the overall quality score and each of the 6 quality dimensions. Word count was included as a fixed-effect covariate, along with an interaction term between responder type (CA vs clinician) and word count. Random intercepts were included for unique response ID and rater to account for clustering of ratings (3 raters per response).

Coefficients represent the estimated change in rating score per additional word. "CA β" and "Clinician β" denote the within-group association between word count and rating score for chatbot- and clinician-authored responses, respectively. The interaction P value tests whether the association between word count and rating differs between CA and clinician responses. Two-sided Wald tests were used to assess statistical significance.

# Multimedia Appendix 7

**S 4: Domain-specific mixed-effects model results comparing CA and clinician responses**

| Domain | Estimated mean difference | 95% CI | P value |
|---|---|---|---|
| A | 1.013 | 0.823 to 1.204 | <0.001 |
| B | 0.722 | 0.555 to 0.889 | <0.001 |
| C | 0.912 | 0.695 to 1.129 | <0.001 |
| D | 0.721 | 0.507 to 0.935 | <0.001 |
| E | 0.419 | 0.053 to 0.785 | 0.0248 |
| F | 0.493 | 0.089 to 0.897 | 0.0168 |

Estimated mean differences (CA – Clinician) in overall quality scores between CA-generated and clinician-generated responses within each predefined content domain, derived from linear mixed-effects models with individual ratings as the outcome. Models included responder type as a fixed effect and random intercepts for case and rater. Values represent CA minus clinician mean differences with 95% CIs and two-sided P values.

# Multimedia Appendix 8

## S 5: Domain- and dimension-specific quality ratings (mean ± SD)

| Domain | Responder | Accuracy | Guidelines | Actionable | Personalized | Clarity | Empathy | Overall Quality |
|---|---|---|---|---|---|---|---|---|
| A | CA | 4.35 ± 0.67 | 4.32 ± 0.65 | 4.42 ± 0.67 | 4.31 ± 0.77 | 4.36 ± 0.68 | 4.34 ± 0.67 | 4.35 ± 0.56 |
| A | Clinician | 3.78 ± 0.97 | 3.69 ± 0.93 | 2.95 ± 1.15 | 3.14 ± 1.19 | 3.53 ± 1.14 | 2.94 ± 1.26 | 3.34 ± 0.94 |
| B | CA | 4.41 ± 0.64 | 4.36 ± 0.62 | 4.52 ± 0.56 | 4.23 ± 0.67 | 4.51 ± 0.57 | 4.36 ± 0.57 | 4.40 ± 0.46 |
| B | Clinician | 3.81 ± 0.95 | 3.81 ± 0.94 | 3.69 ± 1.03 | 3.46 ± 1.05 | 3.86 ± 0.96 | 3.41 ± 1.07 | 3.68 ± 0.88 |
| C | CA | 4.47 ± 0.60 | 4.35 ± 0.77 | 4.38 ± 0.73 | 4.38 ± 0.62 | 4.56 ± 0.58 | 4.38 ± 0.54 | 4.42 ± 0.51 |
| C | Clinician | 3.72 ± 0.98 | 3.78 ± 0.88 | 3.42 ± 1.12 | 3.35 ± 1.16 | 3.60 ± 1.04 | 3.17 ± 1.23 | 3.50 ± 0.86 |
| D | CA | 4.38 ± 0.66 | 4.22 ± 0.79 | 4.45 ± 0.69 | 4.26 ± 0.67 | 4.49 ± 0.58 | 4.54 ± 0.60 | 4.39 ± 0.55 |
| D | Clinician | 3.90 ± 0.92 | 3.82 ± 0.89 | 3.44 ± 1.04 | 3.42 ± 1.06 | 3.78 ± 1.00 | 3.65 ± 1.20 | 3.67 ± 0.86 |
| E | CA | 4.42 ± 0.73 | 4.25 ± 0.73 | 4.28 ± 0.75 | 4.17 ± 0.77 | 4.31 ± 0.67 | 4.31 ± 0.71 | 4.29 ± 0.63 |
| E | Clinician | 4.06 ± 0.95 | 4.14 ± 0.96 | 3.79 ± 1.20 | 3.67 ± 1.17 | 3.97 ± 1.00 | 3.58 ± 1.18 | 3.87 ± 0.93 |
| F | CA | 4.42 ± 1.00 | 4.33 ± 1.01 | 4.35 ± 1.07 | 3.92 ± 1.13 | 4.33 ± 0.96 | 4.14 ± 0.93 | 4.25 ± 0.93 |
| F | Clinician | 3.97 ± 0.84 | 3.97 ± 0.88 | 3.75 ± 1.02 | 3.56 ± 0.94 | 3.89 ± 1.01 | 3.39 ± 0.99 | 3.75 ± 0.84 |

Domain- and dimension-specific quality ratings (mean ± SD) for CA-generated and clinician-generated responses, presented in paired format for each predefined content domain (A–F). Overall quality represents the arithmetic mean of 6 dimension-specific ratings on a 1–5 scale.

# Multimedia Appendix 9

**S 6: Rater-level distribution of overall quality scores stratified by perceived source (clinician vs CA)**

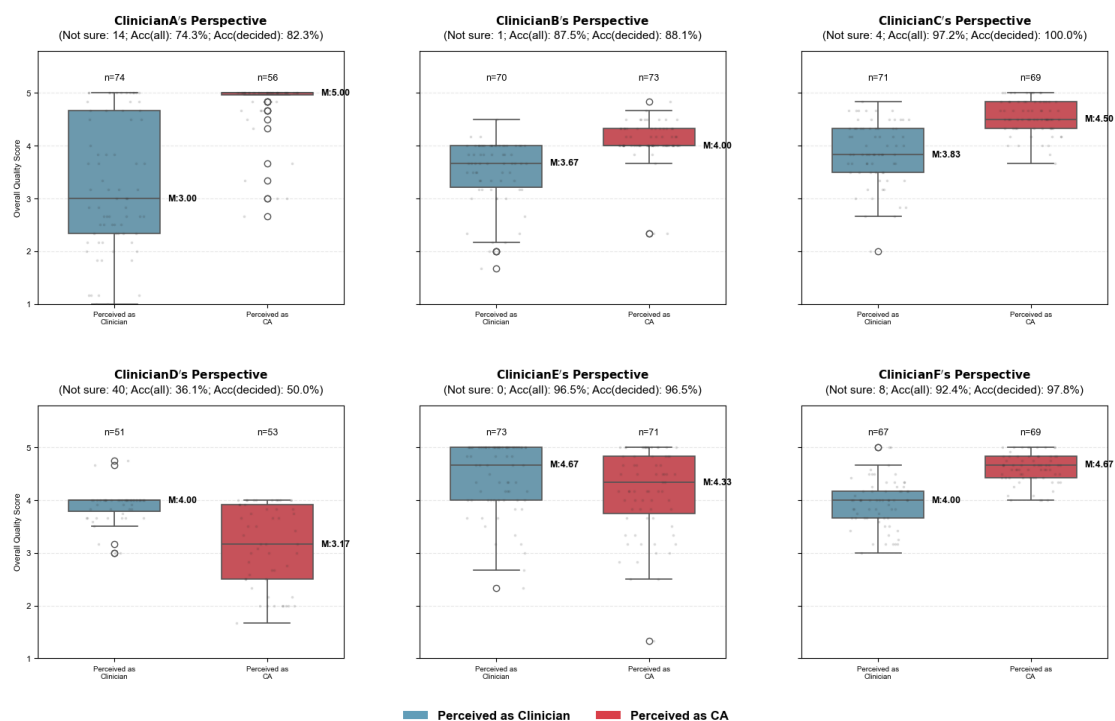

Boxplots show the distribution of overall quality scores assigned by each individual clinician rater (A–F), stratified by the rater's perceived source of the response ("Perceived as Clinician" vs "Perceived as CA").

Overall quality was defined as the arithmetic mean of the 6 5-point quality dimensions (clinical accuracy, guideline adherence, actionability, personalization, clarity, and empathy) for each rating instance. In each panel, M denotes the median overall quality score (not the mean). Boxes represent the IQR, horizontal lines indicate medians, and whiskers extend to 1.5×IQR.

The number of responses in each perceived-source category is shown above each box (n). Responses marked as "Not sure" were excluded from the boxplot stratification but are reported separately in the panel subtitle.

Acc (all) denotes identification accuracy calculated across all rated responses, with "Not sure" classifications treated as incorrect.

Acc (decided) refers to identification accuracy calculated only among responses for which the rater made a definitive source judgment (i.e., excluding "Not sure" responses).

These plots illustrate variability in source discrimination performance across raters and allow visual comparison of the quality scores assigned to responses perceived as clinician-generated versus CA-generated.